\documentclass{article}

\usepackage{PRIMEarxiv}

\usepackage[utf8]{inputenc} 
\usepackage[T1]{fontenc}    
\usepackage{hyperref}       
\usepackage{url}            
\usepackage{booktabs}       
\usepackage{amsfonts}       
\usepackage{nicefrac}       
\usepackage{microtype}      
\usepackage{lipsum}
\usepackage{fancyhdr}       
\usepackage{graphicx}       
\graphicspath{{media/}}     
\usepackage{booktabs}
\usepackage{array}
\usepackage{multirow}
\usepackage{makecell}
\usepackage{float}
\usepackage{pifont}
\usepackage{amsmath}
\usepackage[export]{adjustbox} 
\PassOptionsToPackage{table}{xcolor}
\usepackage{tikz}
\usepackage{colortbl}
\newcommand{\cmark}{\ding{51}}
\newcommand{\xmark}{\ding{55}}

\title{CascadedViT: Cascaded Chunk-FeedForward and Cascaded Group Attention Vision Transformer
}

\author{
  Srivathsan Sivakumar \\
  Ontario Tech University, Canada\\
  \texttt{srivathsan.sivakumar@ontariotechu.net}
   \And
  Faisal Z. Qureshi \\
  Ontario Tech University, Canada\\
  \texttt{Faisal.Qureshi@ontariotechu.ca}
}

\begin{document}
\maketitle

\begin{abstract}
    Vision Transformers (ViTs) have demonstrated remarkable performance across a range of computer vision tasks; however, their high computational, memory, and energy demands hinder deployment on resource-constrained platforms. In this paper, we propose \emph{Cascaded-ViT (CViT)}, a lightweight and compute-efficient vision transformer architecture featuring a novel feedforward network design called \emph{Cascaded-Chunk Feed Forward Network (CCFFN)}. By splitting input features, CCFFN improves parameter and FLOP efficiency without sacrificing accuracy. Experiments on ImageNet-1K show that our \emph{CViT-XL} model achieves 75.5\% Top-1 accuracy while reducing FLOPs by 15\% and energy consumption by 3.3\% compared to EfficientViT-M5.   Across various model sizes, the CViT family consistently exhibits the lowest energy consumption, making it suitable for deployment on battery-constrained devices such as mobile phones and drones. Furthermore, when evaluated using a new metric called \emph{Accuracy-Per-FLOP (APF)}, which quantifies compute efficiency relative to accuracy, CViT models consistently achieve top-ranking efficiency. Particularly, CViT-L is 2.2\% more accurate than EfficientViT-M2 while having comparable APF scores.
\end{abstract}

\section{Introduction}

Vision Transformers (ViTs) are large, sophisticated, and data-hungry models that provide exceptional performance and modeling capacity~\cite{janiesch2021machine, tu2024overview, dosovitskiy2020image, liu2021swin, liu2023efficientvit}. CoCa~\cite{yu2022cocacontrastivecaptionersimagetext}, a 2.1 billion-parameter image-text encoder-decoder foundation model, achieved state-of-the-art 91\% Top-1 classification accuracy on ImageNet-1K~\cite{5206848} using a fine-tuned encoder. In comparison, the best-performing convolutional neural network (CNN) model~\cite{cai2022reversible} achieves 90\% Top-1 classification accuracy on the same dataset.

\begin{figure*}
    \centerline{
    \includegraphics[width=\textwidth]{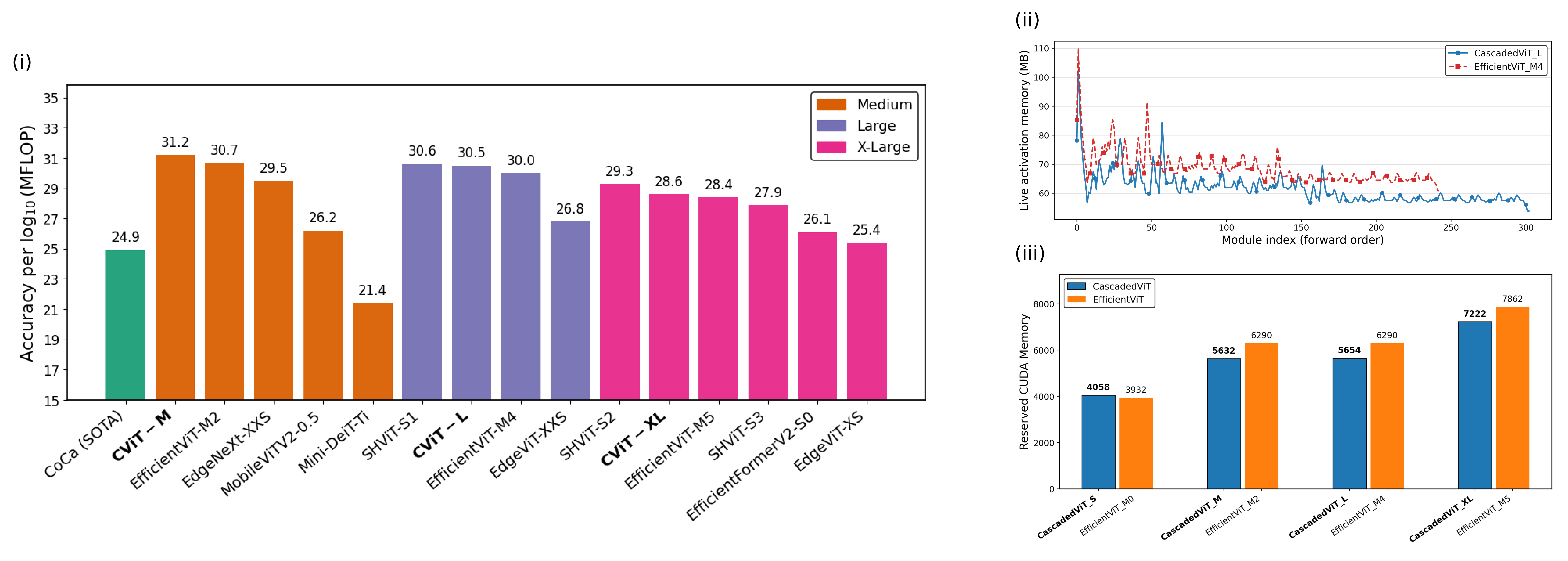}}
    \caption{\textbf{(i)} Accuracy per $\text{log}_{10}$MFLOP (APF) for all evaluated models. Models are grouped by size, and higher APF indicates greater Top-1 accuracy per unit of computation. \textbf{(ii)} Live memory trace of CViT-L and EfficientViT-M4. CViT-L consistently demands less memory indicating system-wide memory efficiency. \textbf{(iii)} Reserved memory (MB) demands of CViT and EfficientViT. CViT demands less memory allocation across all scales except small.}
    \label{fig:apf}
\end{figure*}

Due to their groundbreaking performance, Vision Transformers (ViTs) are increasingly being deployed in real-time applications and battery-constrained devices such as Unmanned Aerial Vehicles (UAVs)~\cite{youn2023compressing, Wu_2025_CVPR, bhattacharya2024vision} and wearable technologies~\cite{robitaille2025vision}. However, models like CoCa~\cite{yu2022cocacontrastivecaptionersimagetext} have substantial computational and memory requirements, making their deployment in resource-limited environments a significant challenge~\cite{NEURIPS2022_5452ad8e}. This highlights the urgent need for the design of lightweight and compute-efficient ViT architectures that can operate effectively under stringent hardware and energy constraints.

Several lightweight Vision Transformer (ViT) architectures have been proposed to reduce model size and complexity for image classification tasks, including MiniViT~\cite{zhang2022minivit}, EdgeViT\cite{pan2022edgevits}, and TinyViT~\cite{tiny_vit}. While these models are considerably smaller than standard ViTs, they still require significantly more computational resources than EfficientViT~\cite{liu2023efficientvit}. As shown in \hyperref[sec:Table2]{Table~2}, the largest EfficientViT variant achieves approximately four times higher throughput than the smallest MiniViT model (Mini-DeiT-Ti). However, most efficient ViT designs primarily optimize for latency, parameter count, and FLOPs, often overlooking energy consumption \cite{liu2023efficientvit, zhang2022minivit, tiny_vit, li2022efficientformer} which is a critical factor for deployment in battery-powered systems.

EfficientViT achieves its efficiency by replacing self-attention modules with feedforward networks (FFNs), which have been shown to exhibit redundancy~\cite{pires2023one}. To mitigate this redundancy and further improve efficiency, techniques such as architecture-adaptive optimization (e.g., weight sharing) and modular designs like Chunk-FFNs (CFFN)~\cite{10651310} have been proposed. These strategies not only reduce memory and energy consumption but may also enhance model accuracy~\cite{zhang2020deeper}.

In this paper, we introduce \textbf{Cascaded-ViT (CViT)}, a ViT variant optimized for battery-powered edge deployment, that addresses the dominant inference bottleneck---the FFNs~\cite{pei2025cmoe}---in EfficientViT by replacing FFNs with Cascaded-Chunk Feed Forward Network (CCFFN).  The CCFFN module partitions the input feature map along the channel dimension into smaller subsets and applies a lightweight FFN independently to each subset. The outputs from earlier FFNs are progressively aggregated through residual connections, enabling deeper feature refinement with minimal overhead. Throughout this work, we use the terms ``EfficientViT'' and ``backbone'' interchangeably to refer to the underlying network framework.

We evaluate CViT against several state-of-the-art efficient and edge-optimized ViT architectures. Experimental results on ImageNet-1K~\cite{5206848} show that CViT achieves competitive accuracy while substantially reducing computational cost, both in terms of energy consumption and memory footprint.
Specifically, \textbf{CViT-M} reduces FLOPs by \textbf{15\%} and parameters by \textbf{0.7M} compared to EfficientViT-M2, with only a \textbf{0.9\%} drop in Top-1 accuracy.  Additionally, CViT yields a smaller memory footprint and lower per-image energy than competing methods.  To better capture the trade-off between accuracy and computational budget, we introduce a new metric---\textbf{Accuracy per FLOP (APF)} which quantifies compute efficiency relative to predictive performance. CViT achieves a favorable position on this APF curve, establishing a new Pareto frontier for real-world ViT deployment under resource constraints.

\section{Related Works}
\label{sec:related_works}

As discussed earlier, Vision Transformer (ViT) models are large and data-intensive, which limits their deployment in resource-constrained environments. Consequently, there is strong motivation to compress these models to enable their use in edge devices and energy-limited systems. Two widely adopted compression strategies include: (1) pruning and (2) weight sharing. However, we argue that efficient architectural design remains the most effective means of achieving ViT efficiency. As demonstrated in \hyperref[tbl:4]{Table~\ref{tbl:ablation}}, compression techniques such as clustered weight sharing often yield limited practical benefits.

Pruning techniques aim to reduce inference latency by removing redundant parameters and optimizing matrix operations~\cite{tang2024survey}. For example, \cite{yang2023global} proposed a Hessian-based pruning strategy that facilitates more effective parameter redistribution. \cite{zheng2022savit} introduced a Taylor-based approximation method to identify strong component matches for collaborative pruning. Token- and feature-level pruning~\cite{kong2022spvit, rao2021dynamicvit, zhu2021vision, tang2022patch} is another common technique that reduces ViT computation by eliminating less informative regions of the input. \cite{10879005} proposed an edge- and energy-aware dynamic pruning method that drops tokens with low attention scores, while retaining those with high scores. \cite{song2021dancing} further demonstrated that dynamic reconfiguration of transformers at runtime, based on hardware and software profiles, can extend battery life by up to 4×.

Weight sharing---first introduced in the 1980s~\cite{lecun1989generalization}---ties parameters across layers to reduce model size and increase efficiency~\cite{zhang2020deeper, kowsher2024does, dehghani2018universal}. While this technique has shown promise in improving parameter efficiency, it can also reduce model flexibility and accuracy if not carefully optimized~\cite{zhang2020deeper}. Recent work has focused on mitigating this trade-off~\cite{ullrich2017soft, DUPUIS2021114148}. MiniViT~\cite{zhang2022minivit}, for instance, introduced weight multiplexing, combining shared weights with transformations to simulate demultiplexing. However, the added transformations increase computational overhead. \cite{wang2024residualtransformer} proposed residual weight sharing, where each shared layer includes a residual component, increasing model capacity with only a modest increase in parameter count.

Efficient architectural design is another promising approach and often involves hybrid CNN-ViT structures~\cite{khan2023survey, wu2021cvt, graham2021levit}. MobileViTv3~\cite{wadekar2022mobilevitv3} replaces local convolutional operations with global transformer-based computations. Its smallest variant, MobileViTv3-XXS, achieves 70.98\% top-1 accuracy on ImageNet-1K~\cite{5206848} with only 1.2 million parameters. EfficientFormer~\cite{li2022efficientformer}, a lightweight transformer architecture, employs latency-aware pruning and maintains consistent tensor dimensions. EfficientFormer-L1 achieves an inference latency of just 1.6 ms on an iPhone 12. Recently, MicroViT~\cite{setyawan2025microvit} introduced an energy-efficient Single Head Attention mechanism that processes only one-quarter of the input features in the attention block, significantly reducing model size and power consumption.

\begin{figure}
    \centerline{
    \includegraphics[width=0.7\linewidth]{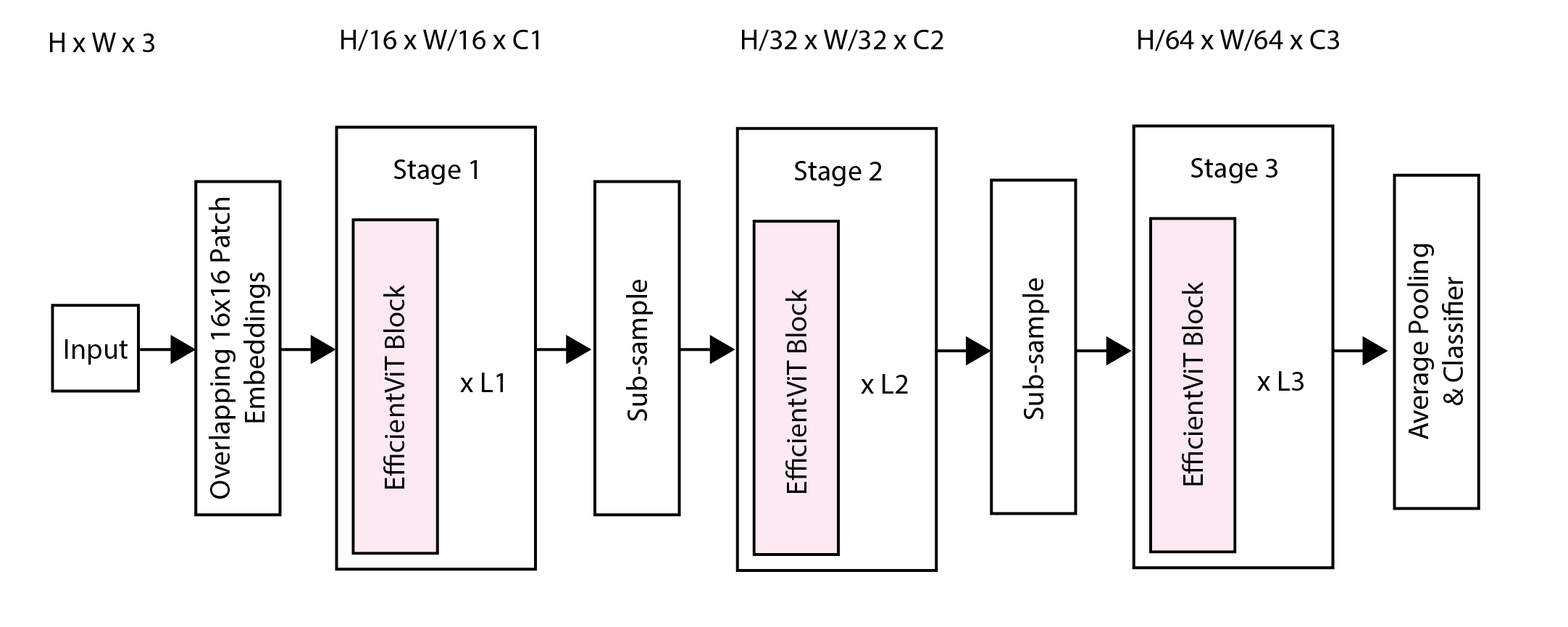}
    }
    \centerline{
    \includegraphics[width=0.7\linewidth]{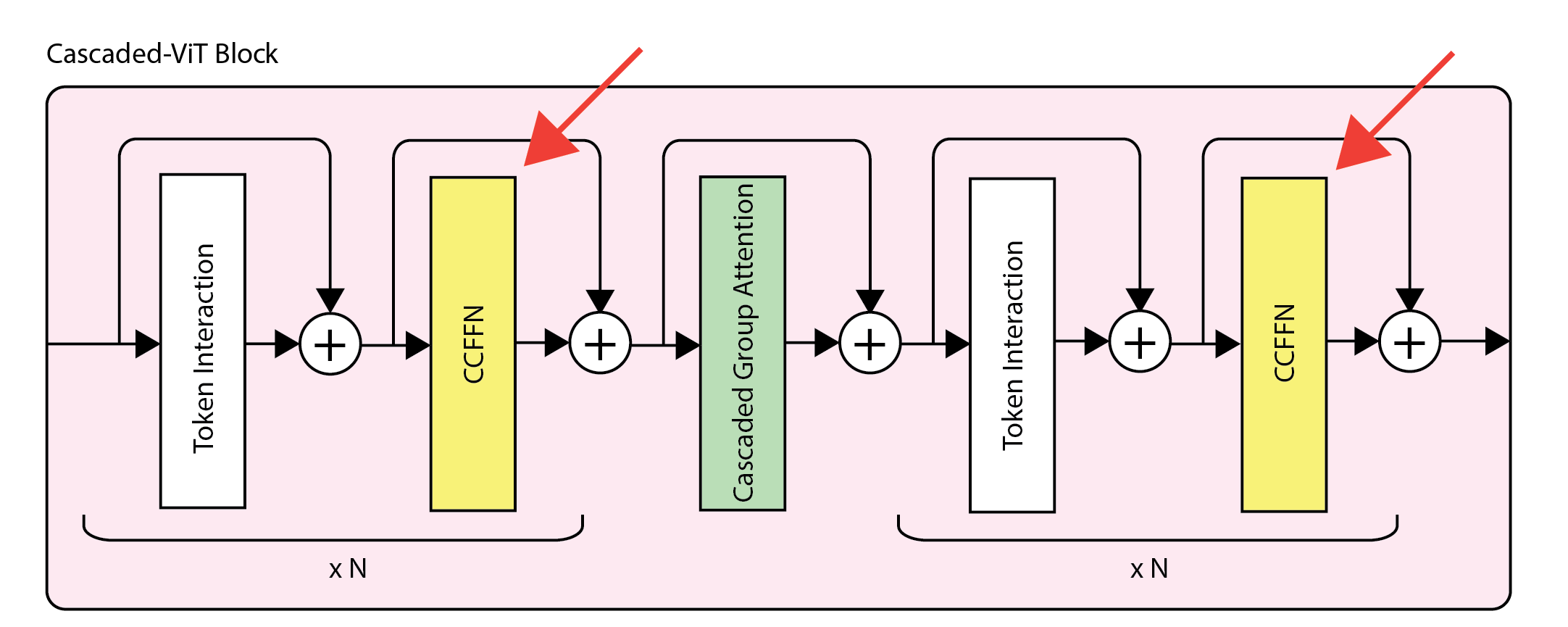}
    }
    \caption{An overview of the EfficientViT (top).  The proposed architecture replaces EfficientViT Block with the Cascaded Chunk ViT Block (bottom).  Cacaded Chunk ViT Block replaces the pre- and post-attention FFNs in the EfficientViT Block with Cascaded Chunk FFNs (shown in Yellow and highlighted with Red arrows).}
    \label{fig:architecture}
\end{figure}

\section{The Making of Cascaded-ViT}
\label{sec:CViT}
Cascaded-ViT replaces the Feed-Forward Network (FFN) layers in EfficientViT with Cascaded Chunked Feed-Forward Network (CCFFN) layers~\cite{liu2023efficientvit} (\hyperref[fig:architecture]{Figure~\ref{fig:architecture}}). We begin by reviewing the architecture of EfficientViT. \hyperref[fig:architecture]{Figure~\ref{fig:architecture}} provides an overview.

At a high level, EfficientViT consists of three stages, each containing a fixed number of EfficientViT blocks. Each block comprises five components: Token Interaction, Pre-Attention Feed-Forward Network, Cascaded Group Attention (CGA), Token Interaction, and Post-Attention Feed-Forward Network.  The input to the first stage is generated by applying overlapping $16 \times 16$ patch embeddings to the input image. Between the stages, subsampling layers reduce the spatial resolution by half. Finally, the output of the last stage is processed via global average pooling and passed to a classifier for prediction.

EfficientViT's efficiency stems from three key architectural innovations:
\begin{enumerate}
\item \textbf{Memory efficiency:} Memory-intensive self-attention layers are replaced with lightweight feed-forward networks (FFNs). A single self-attention module is positioned between pre-attention and post-attention FFNs, significantly reducing memory usage.
\item \textbf{Computational efficiency:} To lower computational cost, only a subset of input features is routed to each attention head. The outputs of successive heads are accumulated and added to the next subset of the input feature, improving performance while reducing FLOPs and parameters.
\item \textbf{Parameter efficiency:} Parameters are strategically reallocated—more to critical components and fewer to less influential ones. Given the high redundancy in FFNs~\cite{pires2023one}, their expansion ratios are halved, leading to a leaner yet effective model.
\end{enumerate}
EfficientViT is particularly well-suited for deployment to edge devices.  For instance, EfficientViT-M0 contains only 2.3 million parameters, achieves 63.2\% top-1 accuracy on ImageNet-1K~\cite{5206848}, and can run inference smoothly on an iPhone 11~\cite{liu2023efficientvit}. 

\subsection{Beyond EfficientViT}

This work began as an investigation into how to make EfficientViT even more ``efficient.'' On one hand, we drew inspiration from model compression techniques such as weight sharing and pruning, which are widely studied for reducing model size and computational cost~\cite{zhang2020deeper, gholami2024can}. On the other hand, we followed principled approaches to efficient network design based on low-complexity Vision Transformers and chunk-level FFN optimization~\cite{10651310, setyawan2025microvit}.

While both directions were explored, we found that replacing the pre- and post-attention FFN layers in EfficientViT blocks with \textit{Cascaded Chunk Feed-Forward Networks (CCFFNs)} led to substantial improvements. Specifically, this modification significantly reduces model complexity in terms of parameter count and FLOPs, while maintaining competitive performance across standard vision tasks such as classification, object detection, and semantic segmentation.  (Results related to weight sharing are included in the ablation study in \hyperref[sec:ablation]{Section~\ref{sec:ablation}}).  Cascaded-ViT uses \textit{Batch Normalization} (BatchNorm)~\cite{ioffe2015batch} throughout the backbone, as it can be fused with convolutional layers to improve inference speed. The activation function used is \textit{ReLU}~\cite{nair2010rectified}, chosen for its computational efficiency and broad hardware compatibility.  Similar to EfficientViT, in each subsampling block, the Cascaded Group Attention (CGA) module is replaced by an \textit{inverted residual block}, following the design principles proposed in~\cite{howard2019searching, sandler2018mobilenetv2}. Each subsampling block also incorporates a \textit{Squeeze-and-Excite (SE)} module~\cite{hu2018squeeze} to enhance feature recalibration and emphasize informative channels, while reducing the spatial resolution by a factor of two.

\begin{table}
\centering
\renewcommand{\arraystretch}{1.0}
\begin{adjustbox}{width=0.6\columnwidth, center}
\tiny
\begin{tabular}{l | c  c  c }
    \toprule
    \textbf{Model} & \textbf{Depth} & \textbf{Emb.\ dim.} & \textbf{\# Heads} \\
    \midrule
    CViT-S & {[}1,\,2,\,3{]}   & {[}64, 128, 192{]} & {[}2, 3, 3{]} \\
    CViT-M & {[}1,\,2,\,3{]}   & {[}128, 192, 224{]} & {[}4, 3, 2{]} \\
    CViT-L & {[}1,\,2,\,3{]}   & {[}128, 256, 384{]} & {[}4, 4, 4{]} \\
    CViT-XL & {[}1,\,3,\,4{]}  & {[}192, 288, 384{]} & {[}3, 3, 4{]} \\
    \bottomrule
  \end{tabular}
\end{adjustbox}
\caption{Cascaded-ViT is available in multiple model sizes. In this configuration, \textit{Depth} refers to the number of Cascaded-ViT blocks in each stage, \textit{Embedding Dimension} indicates the number of feature channels after each stage, and \textit{Heads} denotes the number of attention heads used in the Cascaded Group Attention module.}
\label{tbl:sizes}
\end{table}

\subsection{Chunk FFN}
\label{sec:CFFN}

Originally proposed for a speech recognition Transformer model~\cite{10651310}, the Chunk Feed Forward Network (CFFN) is an efficient compression technique in which the input is partitioned into smaller chunks, each processed by a lightweight feed-forward network (FFN). Following the formulation in~\cite{geva2020transformer}, the FFN is redefined as a key-value mechanism and expressed as:

\newcommand{\bX}{\mathbf{X}}
\newcommand{\bY}{\mathbf{Y}}

\begin{equation}\label{eq:key-valueFFN}
\text{FFN}(\bX) = f(\bX \mathbf{K}^{T}) \mathbf{V}
\end{equation}

where $\mathbf{K}$ and $\mathbf{V}$ denote learnable weight matrices, and $f$ represents a non-linear activation function. The Chunk-FFN effectively captures diverse feature representations while leveraging the key-value memory structure in~\hyperref[eq:key-valueFFN]{Eq.~\ref{eq:key-valueFFN}}, without incurring high computational costs. This design leads to a substantial reduction in parameters while maintaining stable performance. Empirically, the model demonstrated improved results on Aishell-1~\cite{bu2017aishell} and HKUST~\cite{liu2006hkust}, achieving 0.3\% and 0.2\% reductions in Character Error Rate (CER), respectively.


\subsection{Cascaded-Chunk FFN}
\label{sec:CCFFN}

\begin{figure}
    \centerline{
    \includegraphics[width=\linewidth]{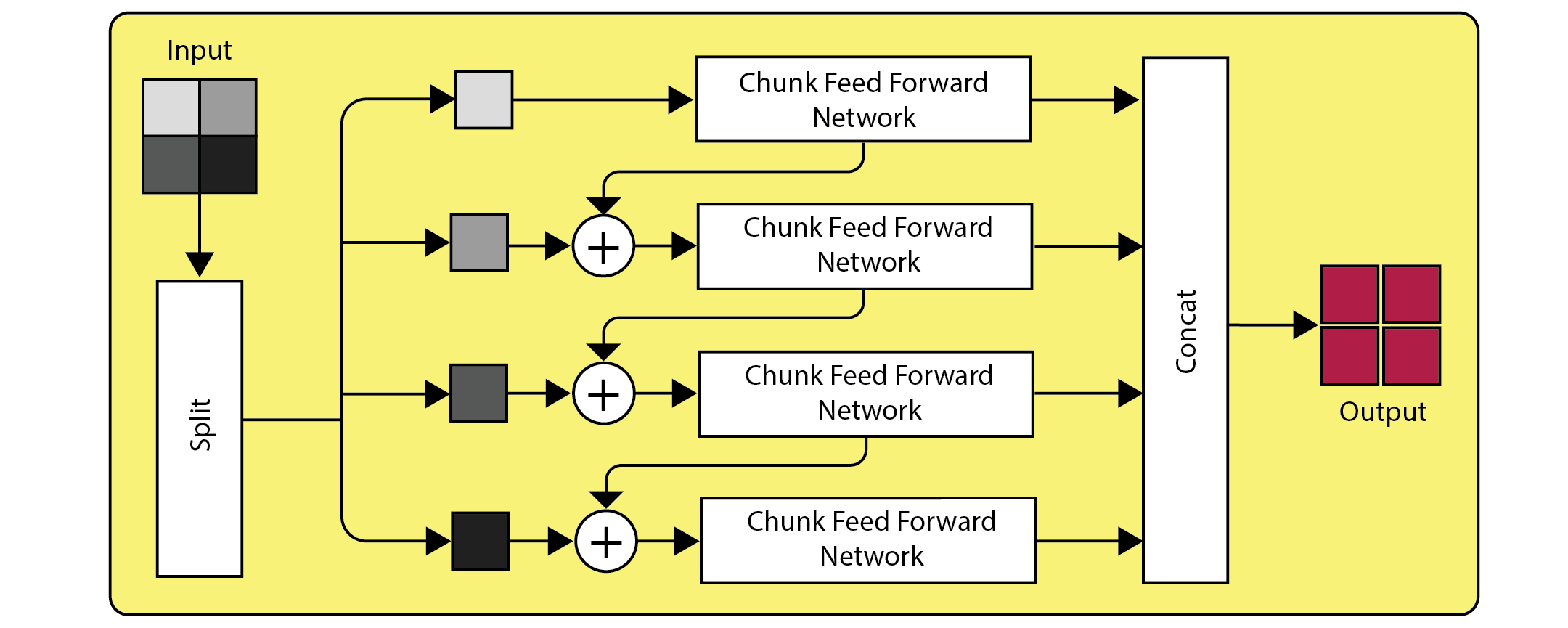}
    }
    \caption{Cascaded-Chunk FFN layer that replaces the FFN layers in EfficientViT Blocks.}
    \label{fig:ccffn}
\end{figure}

Although the expansion ratio of FFN layers in the backbone is reduced from 4 to 2 as in~\cite{liu2023efficientvit}, we argue that these layers can be further optimized to create a smaller and more efficient model. To this end, inspired by CFFN~\cite{10651310} and the Cascaded Group Attention (CGA) mechanism~\cite{liu2023efficientvit} for Vision Transformers, we introduce Cascaded-Chunk FFN (CCFFN).

As illustrated in~\hyperref[fig:ccffn]{Figure~\ref{fig:ccffn}}, the input feature is divided into multiple chunks, each passed through a compact FFN. The output from one FFN is added to the next chunk and then used as the input to the subsequent FFN, forming a cascade of progressively enriched features. Drawing from~\cite{10651310, liu2023efficientvit}, the CCFFN module can be formally expressed as:

\begin{equation}\label{eq:split}
    \bX_{1},\cdots,\bX_{n} = \text{Split}(\bX),
\end{equation}
\begin{equation}\label{eq:cascading}
\bX'_i = 
    \begin{cases}
        \bX_i, & \text{if } i = 1 \\
        \bX_i + Y_{i-1}, & \text{if } i > 1
    \end{cases}
\end{equation}
\begin{equation}\label{eq:chunkFFN}
    \bY_{i} = \text{FFN}_{i}(\bX'_{i}),
\end{equation}
\begin{equation}\label{eq:Concat}
    \text{Output} = \text{Concat}(\bY_{1},\cdots,\bY_{n}),
\end{equation}

where $1 \le i \le n$ and $n$ is the number of chunks. $\bX_i$ is the $i^{th}$ subset of the input. $\bX_{i}'$ is the input to the FFN. $\bY_i$ is the $i^{th}$ output from the CFFN. We further boost the capacity of the model by computing the feature maps of each CFFN in a cascaded manner as defined in \hyperref[eq:cascading]{Eq 3}. The output from each CFFN is added to the next subset of input which promotes the learning of more refined features. $\bX'_{i}$ is the sum of the $i^{th}$ subset of the input $\bX_{i}$ and the $(i-1)^{th}$ output from the CFFN. When $i = 1$, the input remains unaffected as there are no CFFNs before the first one.

This new CCFFN module saves approximately 20{\%} parameters and 15{\%} FLOPs compared to the backbone.  Crucially, CCFFN posts lower live and reserved memory demands, which is highly desirable for deployment in resource-constrained edge devices~\cite{naveen2022memory, cheng2023memory, maas2022telamalloc}.  Furthermore, The cascading nature of the module increases the depth of the model, potentially boosting its capacity while not adding to the parameter count. Since we are launching more FFNs compared to the backbone, a slight but noticeable increase in latency is expected.

\section{Experiments}
The training recipe used by \cite{liu2023efficientvit} introduces instability, likely due to aggressive image augmentations being incompatible with FFNs that operate on reduced channel dimensions. As a result, careful hyperparameter tuning is necessary to stabilize training under the modified architecture. We train CViT from scratch on ImageNet-1K~\cite{5206848} using a single NVIDIA GH200-96GB GPU for 300 epochs, employing the AdamW optimizer~\cite{loshchilov2017decoupled}. The weight decay is set to $1.25 \times 10^{-2}$, with a maximum learning rate of $9 \times 10^{-4}$ and a minimum of $1 \times 10^{-4}$. A batch size of 3072 is used to further reinforce training stability. Mixup and CutMix ratios are set to 0.6 and 0.8, respectively.

We adopt the same augmentation techniques and single-cycle cosine annealing schedule as in \cite{liu2023efficientvit, touvron2021training}, which include AutoAugment~\cite{Cubuk_2019_CVPR}, Mixup~\cite{zhang2017mixup}, and Random Erasing~\cite{zhong2020random}. 

To evaluate computational performance, we collect throughput metrics across different platforms using a batch size of 2048. GPU inference is measured on an NVIDIA V100-SXM2-32GB, Apple Silicon performance on an Apple M4 Pro chip, and NPU inference on an AMD Ryzen AI chip. For Ryzen AI, models are quantized to 8-bit integers and converted to ONNX format for compatibility. CPU throughput is measured on an Intel Xeon Silver 4114 @ 2.20GHz with batch size 16, using a single thread in accordance with \cite{graham2021levit}.

Energy consumption during inference is measured on a MacBook Pro with the Apple M4 Pro chip. GPU power readings are obtained via \texttt{powermetrics} at 1 Hz sampling frequency, which reports two GPU readings per second. The idle GPU power range is identified as $(8 \pm 1) \times 10^{-3}$ W. During inference, peak GPU power values are recorded and averaged across approximately 133 readings per model. With each model running for $133/2 = 66.5$ seconds on average, total energy consumption is computed as the product of average power usage and runtime duration.

\subsection{Knowledge Distillation}

We train CViT-XL and CViT-L from scratch on ImageNet-1K~\cite{5206848} using the training recipe described previously, and employ them as teacher models in a knowledge distillation (KD) framework to enhance the performance of smaller models. During student training, we reduce the CutMix and Mixup ratios to 0.1, set the distillation loss weight $\alpha$ to 0.5, and use a temperature of 2.0. Additionally, we lower the minimum learning rate to $9 \times 10^{-5}$ to facilitate finer weight updates during later training stages.

Interestingly, using CViT-XL as the teacher for CViT-S and CViT-M led to a performance drop, even when experimenting with higher temperature values to soften the teacher's outputs. This degradation is likely due to the teacher's predictions being overly confident, which small-capacity student models struggle to learn from effectively~\cite{huang2022knowledge}. Consequently, we selected CViT-L as the teacher for CViT-S and CViT-M, and used CViT-XL to distill CViT-L, keeping the same KD hyperparameters across all configurations.

The results show consistent gains: CViT-S, CViT-M, and CViT-L improve their Top-1 accuracy by 2\%, 1.4\%, and 0.6\%, respectively, when trained with their corresponding teacher models.

\subsection{ImageNet-1K}

We compare our CViT models against several small and lightweight Vision Transformers, including EfficientViT~\cite{liu2023efficientvit}, MiniViT~\cite{zhang2022minivit}, TinyViT~\cite{tiny_vit}, SHViT~\cite{Yun_2024_CVPR}, and EdgeViT~\cite{pan2022edgevits}, with results summarized in \hyperref[sec:Table2]{Table~2}.  CViT consistently demonstrates the lowest or near-lowest computational and energy requirements while maintaining competitive Top-1 accuracy. The integration of the CCFFN module enables our models to remain among the smallest within each size category.

While some models such as EfficientFormerV2-S0~\cite{li2022efficientformer} contain fewer parameters, they incur substantially higher FLOPs (e.g., 800 MFLOPs). In contrast CViT achieves both parameter and compute efficiency, offering a 20\% reduction in parameter count and a 15\% reduction in FLOPs compared to the baseline backbone across all model sizes.
CViT-S is particularly efficient, requiring the least FLOPs and energy among its peers, with only a 1.3\% Top-1 accuracy trade-off compared to EfficientViT-M0~\cite{liu2023efficientvit}. Compared to MobileViTv2-0.5~\cite{mehta2022separable}, CViT-M uses \(2.8 \times \) less compute while maintaining similar accuracy. TinyViT-5M~\cite{tiny_vit} achieves 9.2\% higher accuracy than CViT-M but at a cost of \(15 \times \) more FLOPs.

SHViT-S1~\cite{Yun_2024_CVPR} and CViT-L are comparable in accuracy, with CViT-L achieving a 0.2\% higher Top-1 score at the expense of just 8 additional MFLOPs. Notably, SHViT-S1 consumes \(1.25 \times \) more energy during inference, indicating that CViT-L, despite slightly higher compute, is more energy-efficient. CViT-L, while two sizes larger than EfficientViT-M2, demands similar energy consumption while being 2.2\% more accurate. A similar trend is observed between SHViT-S2 and CViT-XL, where CViT-XL achieves 0.3\% higher accuracy, has 1.6M fewer parameters, only ~16\% more FLOPs and consumes nearly 120mJ/img less energy during inference.
CViT-XL exhibits FLOP counts that are orders of magnitude lower than models like TinyViT~\cite{tiny_vit}, MiniViT~\cite{zhang2022minivit}, and EdgeViT~\cite{pan2022edgevits}, despite incurring only a 5.4\%–5.8\% accuracy trade-off. Across all sizes, CViT requires an average of 5\% less energy compared to EfficientViT.

CViT’s accuracy/FLOP gains are modest, but its energy and memory savings are achieved over an already highly optimized SOTA baseline, so they’re non-trivial. By directly targeting edge bottlenecks, it is expected that CViT will deliver longer battery life or higher throughput, making it a strong deployment choice despite small accuracy deltas.  These results render CViT an attractive solution for deployment in battery-powered and compute-constrained environments where a balance between efficiency and performance is critical.

\begin{table*}
\label{sec:Table2}
    \centering
    \Large
    \begin{adjustbox}{max width=\textwidth}
    \begin{tabular}{l | c c | c c c c | c c c c | c }
         \toprule
         \multirow{2}{*}{\textbf{Model}} &
         \multirow{2}{*}{\makecell[c]{\textbf{Top-1}\\(\%)}} &
         \multirow{2}{*}{\makecell[c]{\textbf{Top-5}\\(\%)}} &
         \multirow{2}{*}{\makecell[c]{\textbf{FLOPs}\\(M)}} &
         \multirow{2}{*}{\makecell[c]{\textbf{Params}\\(M)}} &
         \multirow{2}{*}{\textbf{Input}} &
         \multirow{2}{*}{\textbf{Epochs}} &
         \multicolumn{4}{c|}{\textbf{Throughput} (images/s)} &
         \multirow{2}{*}{\makecell[c]{\textbf{Energy/img}\\(mJ)}} \\
         \cline{8-11}
         & & & & & & & GPU & CPU & M4 Pro & RyzenAI &  \\
         \midrule
        \rowcolor{gray!20}
         \textbf{CViT-S} & 62.0 & 84.2 & \textbf{67} & 1.9 & 224 & 300 & 25740 & \textbf{107} & 5775 & 1453 & $\mathbf{471\pm29}$\\
         EfficientViT-M0 \cite{liu2023efficientvit} & 63.2 & 85.4 & 79 & 2.3 & 224 & 300 & \textbf{30692} & 107 & \textbf{5931} & \textbf{1590} & $490\pm23$ \\
         EdgeNeXt-XXS \cite{maaz2022edgenext} & 71.2 & - & 261 & \textbf{1.3} & 256 & 300 & - & 32 & 930 & - & $\mathit{1166\pm35}$ \\
         MobileViTV2-0.5 \cite{mehta2022separable} & 70.2 & - & 480 & 1.4 & 256 & 300 & 4403 & 9 & 24 & 17 & - \\
        Mini-DeiT-Ti \cite{zhang2022minivit} & \textbf{73.0} & 91.6 & 2600 & 3.0 & 224 & 300 & 3145 & 4 & 731 & 96 & $714\pm39$ \\
        \midrule
         \rowcolor{gray!20}
         \textbf{CViT-M} & 69.9 & 89.5 & \textbf{173} & \textbf{3.5} & 224 & 300 & 20464 & \textbf{58} & 3717 & 867 & $\mathbf{568\pm52}$\\
         EfficientViT-M2 \cite{liu2023efficientvit} & 70.8 & 90.2 & 201 & 4.2 & 224 & 300 & \textbf{21432} & 58 & \textbf{3738} & \textbf{911} & $581\pm36$ \\
         EdgeViT-XXS \cite{pan2022edgevits} & 74.4 & - & 600 & 4.1 & 224 & 300 & 4513 & 15 & 1056 & 119 & $665\pm19$\\
         EfficientFormerV2-S0 \cite{li2022efficientformer} & 75.7 & - & 800 & 3.5 & 224 & 300 & 1180 & 16 & 44 & 185 & $\mathit{1286\pm32}$ \\
         TinyViT-5M \cite{tiny_vit} & \textbf{79.1} & 94.8 & 2600 & 5.4 & 224 & 300 & 3148 & 6 & 96 & - & $\mathit{1170\pm87}$ \\
        \midrule
         SHViT-S1 \cite{Yun_2024_CVPR} & 72.8 & 91.0 & \textbf{241} & \textbf{6.3} & 224 & 300 & \textbf{23132} & \textbf{63} & \textbf{4513} & \textbf{1168} & $737\pm16$\\
         \rowcolor{gray!20}
         \textbf{CViT-L} & 73.0 & 91.2 & 249 & 7.0 & 224 & 300 & 17335 & 45 & 2978 & 667 & $\mathbf{588\pm42}$\\
         EfficientViT-M4 \cite{liu2023efficientvit} & 74.3 & 91.8 & 299 & 8.8 & 224 & 300 & 18132 & 42 & 2894 & 742 & $620\pm45$ \\
         EdgeViT-XS \cite{pan2022edgevits} & \textbf{77.2} & - & 1100 & 6.7  & 224 & 300 & 3502 & 9 & 472 & 28 & $\mathit{1363\pm64}$ \\
        \midrule
         SHViT-S2 \cite{Yun_2024_CVPR} & 75.2 & 92.4 & \textbf{366} & 11.4 & 224 & 300 & 7738 & \textbf{44} & \textbf{2903} & \textbf{904} & $783\pm16$\\
         \rowcolor{gray!20}
         \textbf{CViT-XL} & 75.5 & 92.4 & 435 & \textbf{9.8} & 224 & 300 & 11934 & 27 & 1910 & 423 & $\mathbf{653\pm16}$ \\
         EfficientViT-M5 \cite{liu2023efficientvit} & 77.1 & 93.4 & 522 & 12.4 & 224 & 300 & \textbf{12098} & 26 & 1900 & 556 & $675\pm23$\\
         EdgeViT-S \cite{pan2022edgevits} & 81.0 & - & 1900 & 11.1 & 224 & 300 & 2455 & 6 & 38 & 19 & $\mathit{1379\pm73}$ \\
         TinyViT-11M \cite{tiny_vit} & \textbf{81.5} & 95.8 & 4000 & 11.0 & 224 & 300 & 2509 & 4 & 69 & - & $\mathit{1202\pm191}$ \\
         Mini-DeiT-S \cite{zhang2022minivit} & 80.9 & 95.6 & 9400 & 11.0 & 224 & 300 & 1636 & 2 & 280 & 24 & $773\pm16$\\   
        \bottomrule 
    \end{tabular}
    \end{adjustbox}
    \caption{CViT performance on ImageNet-1K \cite{5206848} compared to state-of-the-art lightweight and efficient ViT models. Throughput is tested on Nvidia V100-SXM2-32GB GPU, Intel Xeon Silver 4114 CPU @ 2.20GHz processor for CPU, Apple M4 Pro GPU for M4 Pro and AMD NPU for RyzenAI. Higher throughput indicates faster inference. Models are not traced before calculating throughput. FLOPs are calculated with fvcore \cite{fvcore2019}. Models are grouped by parameters and sorted in ascending order of FLOPs. Energy values in italic font indicate that a batch size of 1024 was used instead of 2048.  We include Mini-DeiT-Ti and TinyViT-5M---even though they are slightly larger than the target parameter range---since their architectures pursue CViT-like objectives.}
    \label{tbl:classification}
\end{table*}

\subsection{Transfer Learning}

We inspect the transferability of CViT by applying it to the COCO dataset \cite{lin2014microsoft} using an NVIDIA A100-40GB GPU for object detection and segmentation.

\subsubsection{Object Detection}

\hyperref[tbl:detection]{Table~\ref{tbl:detection}} presents the transfer learning results for object detection on COCO~\cite{lin2014microsoft} dataset. Following the setup of EfficientViT~\cite{liu2023efficientvit}, we set the learning rate to 1.5e-4, reduce the weight decay to 0.025, and use a batch size of 32. Under these conditions, CViT-L achieves an average precision (AP) of 29.5 with only 7M parameters, demonstrating its compactness and competitiveness. In comparison, EfficientViT achieves 3.2 AP higher but comes with a 1.8M parameter overhead and increased computational requirements. MicroViT-S3 performs better still, with 36.0 AP, but its 26.7M parameters result in substantial memory demands, making it less suitable for deployment in constrained environments.

CViT-L thus offers a compelling trade-off between accuracy, model size, and efficiency, making it an attractive option for edge applications. Notably, when trained on a 2$\times$ schedule, CViT-L improves to 31.4 AP, highlighting its capacity for further performance gains with extended training.

\begin{table}
  \centering
  \begin{adjustbox}{width=0.7\columnwidth, center}
      \begin{tabular}{l | *{6}{c} | c | c}
        \toprule
        \multirow{2}{*}{\textbf{Model}} &
        \multicolumn{6}{c|}{RetinaNet 1x} & Flops & Params \\
        & AP & AP$_{50}$ & AP$_{75}$ & AP$_{S}$ & AP$_{M}$ & AP$_{L}$ & (M) & (M) \\
        \midrule
        \textbf{CViT-L} & 29.5 & 48.4 & 30.5 & 15.7 & 31.1 & 40.5 & 249 & \textbf{7.0} \\
        EfficientViT-M4 \cite{liu2023efficientvit} & 32.7 & 52.2 & 34.1 & 17.6 & 35.3 & 46.0 & 299 & 8.8 \\
        MicroViT-S3 \cite{setyawan2025microvit} & 36.0 & 56.6 & 38.2 & - & - & - & \textbf{159} & 26.7 \\
        PVT-Tiny \cite{wang2021pyramid} & 36.7 &  56.9 &  38.9 & 22.6 & 38.8 & 50.0 & 1900 & 13.2 \\
        EdgeViT-XXS \cite{pan2022edgevits} & \textbf{38.7} & \textbf{59.0} & \textbf{41.0} & \textbf{22.4} & \textbf{42.0} & \textbf{51.6} & 600 & 13.1 \\ 
        \midrule
        \textbf{CViT-L (2x)} & 31.4 & 50.7 & 32.6 & 16.6 & 33.5 & 44.0 & 249 & 7.0 \\
        \bottomrule
  \end{tabular}
\end{adjustbox}
\caption{Transfer learning results on COCO \cite{lin2014microsoft} object detection using a RetinaNet~\cite{lin2017focal} 1x schedule.}
\label{tbl:detection}
\end{table}

\subsubsection{Instance Segmentation}

\hyperref[tbl:segmentation]{Table~\ref{tbl:segmentation}} presents the transfer learning results on COCO~\cite{lin2014microsoft} instance segmentation using Mask R-CNN~\cite{he2017mask} with a 1$\times$ schedule. We adopt the same training configuration used for object detection, except for a reduced learning rate of 1.25e-4 (down from 1.5e-4). While models like PVT-Tiny~\cite{wang2021pyramid} and EdgeViT-XXS~\cite{pan2022edgevits} incur additional parameter overhead when adapted for segmentation, our CViT-L maintains a fixed parameter count across tasks. This makes it a more compact and resource-efficient choice, while still delivering competitive performance. Additionally, given the performance gains observed with extended training in object detection, we expect further improvements in AP scores for CViT-L under a 2$\times$ schedule.

\begin{table}[t]
  \centering
  \begin{adjustbox}{width=0.7\columnwidth, center}
      \begin{tabular}{l | *{6}{c} | c | c}
        \toprule
        \multirow{2}{*}{\textbf{Model}} &
        \multicolumn{6}{c|}{Mask R-CNN 1$\times$} & Flops & Params \\
        & AP$^b$ & AP$^b_{50}$ & AP$^b_{75}$ & AP$^m$ & AP$^m_{50}$ & AP$^m_{75}$ & (M) & (M) \\
        \midrule
        \rowcolor{gray!20}
        \textbf{CViT-L} & 29.1 & 50.2 & 29.7 & 28.0 & 47.2 & 29.1 & 249 & \textbf{7.0} \\
        EfficientViT-M4 \cite{liu2023efficientvit} & 32.8 & 54.4 & 34.5 & 31.0 & 51.2 & 32.2 & 299 & 8.8 \\
        LightViT-T \cite{huang2022lightvit} & 37.8 & 60.7 & 40.4 & 35.9 & 57.8 & 38.0 & - & 28 \\ 
        PVT-Tiny \cite{wang2021pyramid} & 36.7 &  59.2 &  39.3 & 35.1 & 56.7 & 37.3 & - & 32.9 \\
        EdgeViT-XXS \cite{pan2022edgevits} & \textbf{39.9} & \textbf{62.0} & \textbf{43.1} & \textbf{36.9} & \textbf{59.0} & \textbf{39.4} & - & 23.8 \\ 
        \bottomrule
  \end{tabular}
\end{adjustbox}
\caption{Transfer learning results on COCO \cite{lin2014microsoft} instance segmentation using a Mask R-CNN 1x schedule.}
\label{tbl:segmentation}
\end{table}

\subsection{Latency on iPhone 15 Pro}

We measure the per-image inference latency of our CViT models, along with several state-of-the-art baselines, on an Apple iPhone 15 Pro (See Table~\ref{tbl:latencies}). All measurements are obtained using Apple’s Xcode Instruments profiler to ensure accurate timing under realistic mobile deployment conditions.  The results reported below provide a fair comparison of mobile-side efficiency, highlighting how CViT balances accuracy with responsiveness on a modern smartphone.

We note that CViT consistently outperforms the backbone. Particularly, CViT-XL is 15\% faster than EfficientViT-M5. While the improvements over the backbone may seem incremental for other scales (CViT-S, CViT-M and CViT-L), we emphasize that these are per image latencies. Small reductions at this level accumulate significantly, directly translating to longer battery life or faster inferences when models are deployed at scale or used for prolonged sessions. 

We consider the comparison between CViT-S and EfficientViT-M0 to further illustrate the impact of the gains of our model. The latency gap between CViT-S and EfficientViT-M0 is a mere 0.03 ms per image. However, this translates to roughly 30 seconds of saved compute per million inferences. In the case of CViT-XL it is 130 seconds or 2 minutes and 10 seconds of saved compute. These gains translate to better energy efficiency which is critical for battery-powered edge-deployment.

CViT-based models also demonstrate favorable latency–accuracy trade-offs compared to other leading architectures. For instance, CViT-M matches the accuracy of MobileViTV2-0.5 while offering a faster inference time by 0.15 ms per image. CViT-XL attains accuracy on par with EfficientFormerV2-S0 but delivers a substantial 30\% reduction in latency. Moreover, it surpasses SHViT-S2 by 0.3\% in accuracy and reduces model size by 1.6M parameters, all while maintaining identical latency. Taken together, these results underscore that CViT not only achieves meaningful efficiency gains but also delivers improvements in accuracy and compactness. Such characteristics make CViT especially appealing for mobile and edge deployments, where even millisecond-level latency reductions compound into tangible performance and energy savings.

\label{sec:Table1}
\begin{table}[h]
    \centering
    \begin{adjustbox}{width=0.6\columnwidth, center}
        \begin{tabular}{l | c | c | c }
        \toprule
             \textbf{Model} & 
             \makecell[c]{\textbf{Top-1}\\({\%})} & 
             \makecell[c]{\textbf{Params}\\(M)} &
             \makecell[c]{\textbf{Latency}\\(ms/image)} \\
             \midrule
             \rowcolor{gray!10}
             \textbf{CViT-S} & 62.0 & 1.9 & \textbf{0.39} \\
             EfficientViT-M0 \cite{liu2023efficientvit} & \textbf{63.2} & 2.3 & 0.42 \\
             EdgeNeXt-XXS \cite{maaz2022edgenext} & 71.2 & \textbf{1.3} & 0.51 \\
             MobileViTV2-0.5 \cite{mehta2022separable} & 70.2 & 1.4 & 0.60 \\
             \midrule
             \rowcolor{gray!10}
             \textbf{CViT-M} & 69.9 & \textbf{3.5} & \textbf{0.45} \\
             EfficientViT-M2 \cite{liu2023efficientvit} & 70.8 & 4.2 & 0.48 \\
             EdgeViT-XXS \cite{pan2022edgevits} & 74.4 & 4.1 & 0.80 \\
             EfficientFormerV2-S0 \cite{li2022efficientformer} & \textbf{75.7} & 3.5 & 1.22 \\
             \midrule
             SHViT-S1 \cite{Yun_2024_CVPR} & 72.8 & \textbf{6.3} & \textbf{0.58} \\
             \rowcolor{gray!10}
             \textbf{CViT-L} & 73.0 & 7.0 & 0.70 \\
             EfficientViT-M4 \cite{liu2023efficientvit} & 74.3 & 8.8 & 0.79 \\
             EdgeViT-XS \cite{pan2022edgevits} & \textbf{77.2} & 6.7 & 1.25 \\
             \midrule
             SHViT-S2 \cite{Yun_2024_CVPR} & 75.2 & 11.4 & 0.86 \\
             \rowcolor{gray!10}
             \textbf{CViT-XL} & 75.5 & \textbf{9.8} & \textbf{0.86} \\
             EfficientViT-M5 \cite{liu2023efficientvit} & \textbf{77.1} & 12.4 & 0.99 \\
             \bottomrule
        \end{tabular}
    \end{adjustbox}
    \caption{Measured latency comparison of CViT and other efficient architectures on Apple iPhone 15 Pro. Latency is collected using XCode Instruments (Version 16.2) with CoreML.}
    \label{tbl:latencies}
\end{table}
      
\subsection{Accuracy Per Flop}

\cite{schwartz2020green} advocates for increased research focus on resource- and energy-efficient AI practices, such as using smaller datasets and designing computationally efficient models. This call is particularly relevant given the prevailing trend in model development that prioritizes accuracy and scale, often at the cost of significant energy consumption. For example, \cite{barbierato2024toward} estimates the training cost of GPT-3~\cite{brown2020language} to be approximately 50 MWh, which corresponds to the lifetime carbon emissions of five gasoline-powered cars. These concerns underscore the importance of evaluating model efficiency not solely in terms of predictive accuracy but also in relation to computational cost.

To address this, we propose a new metric aligned with the principles of ``Green AI''~\cite{schwartz2020green}, termed \textbf{Accuracy Per FLOP (APF)}, inspired by \cite{Esposito_2024}. APF measures the accuracy gained per Mega-FLOP (MFLOP), offering a quantifiable indicator of a model’s computational efficiency. A higher APF indicates a model delivers more predictive performance per unit of compute, while a lower APF suggests inefficiency.
We note from several studies that the relationship between accuracy and FLOPs is non-linear~\cite{tay2021scale, tan2019efficientnet, kaplan2020scaling} where accuracy tends to saturate with increasing compute or FLOPs.  Therefore, it is necessary to design a metric that does not assume a linear accuracy-FLOP relationship and thereby favor models that achieve superficial accuracy with very low compute.

To address this we first apply an accuracy threshold ($\ge$70\% Top-1 on ImageNet) following \cite{yang2024double} to ensure that only models that achieve meaningful accuracy are included in comparative studies. In contrast to \cite{yang2024double} where the energy term is anchored to the model with the highest energy, we aim to eliminate APF's dependence on any model. Therefore, we apply log-scaling to the denominator---the log-scaled FLOP term grows more slowly in metric as raw FLOP count increases, capturing the relationship between accuracy and FLOPs observed in~\cite{tay2021scale, tan2019efficientnet, kaplan2020scaling}. This results in a model-agnostic normalization that more faithfully reflects the non-linear relationship between accuracy and FLOPs. 
APF highlight models that achieve high accuracy with efficient compute without disproportionately penalizing larger models that necessarily operate in the high-compute, diminishing-return regime, as would be the case with a linear formulation. Formally APF is defined as
\begin{equation}
    \text{APF} \;=\; \frac{\text{Top-1 Accuracy (\%)}}{\log_{10}\!\text{MFLOP}}.
\end{equation}

Our CViT family consistently ranks among the most compute efficient models. We include CViT-M in this study as its 69.9\% accuracy is $\approx$ 70\% given the $\pm$ 0.1\% variation from seeds and data shuffling, and it matches the performance of MobileViTv2-0.5~\cite{mehta2022separable} on ImageNet-1K~\cite{5206848} while being 15\% more efficient. CViT-L attains an APF of 30.5 with accuracy comparable to SHViT-S1 and to EfficientViT-M2 at a similar APF.  CViT-XL and EfficientFormerV2-S0 achieve similar accuracy; however, CViT-XL yields 10\% higher APF, indicating a stronger accuracy-compute trade-off.  Across sizes, CViT delivers $\sim$ 2\% higher Top-1 than its EfficientViT backbone at comparable compute.

\begin{table}[t]
    \centering
    \begin{adjustbox}{width=0.55\columnwidth, center}
    \tiny
        \begin{tabular}{l | c | c | c}
        \toprule
            \textbf{Model} & 
            \makecell[c]{\textbf{Top-1}\\({\%})} & 
            \makecell[c]{\textbf{FLOPs}\\(M)} &
            \makecell[c]{\textbf{APF}} \\
        \midrule
            CoCa (SOTA) \cite{yu2022cocacontrastivecaptionersimagetext} & 91.0 & 4540 & 24.9 \\
        \midrule
            \rowcolor{gray!20}
            CVIT-M                    & 69.9 & \textbf{173}  & \textbf{31.2} \\
            EfficientViT-M2           & 70.8 & 201           & 30.7 \\
            MobileViTV2-0.5           & 70.2 & 480           & 26.2 \\
            EdgeNeXt-XXS              & 71.2 & 261           & 29.5 \\
            Mini-DeiT-Ti              & \textbf{73.0} & 2600 & 21.4 \\
        \midrule
            SHViT-S1                  & 72.8 & \textbf{241}  & \textbf{30.6} \\
            \rowcolor{gray!20}
            CVIT-L                    & 73.0 & 249           & 30.5 \\
            EfficientViT-M4           & 74.3 & 299           & 30.0 \\
            EdgeViT-XXS               & 74.4 & 600           & 26.8 \\
        \midrule
            SHViT-S2                  & 75.2 & \textbf{366}  & \textbf{29.3} \\
            \rowcolor{gray!20}
            CVIT-XL                   & 75.5 & 435           & 28.6 \\
            EfficientFormerV2-S0      & 75.7 & 800           & 26.1 \\
            EfficientViT-M5           & 77.1 & 522           & 28.4 \\
            SHViT-S3                  & 77.4 & 601           & 27.9 \\
            EdgeViT-XS                & \textbf{77.2} & 1100 & 25.4 \\
        \bottomrule
        \end{tabular}
    \end{adjustbox}
    \caption{Comparison of Top-1 accuracy, FLOPs, and Accuracy Per FLOP (APF) for all evaluated models where larger APF means greater efficiency.}
    \label{tbl:apf}
\end{table}

\begin{figure}
    \centerline{
    \includegraphics[width=0.8\linewidth]{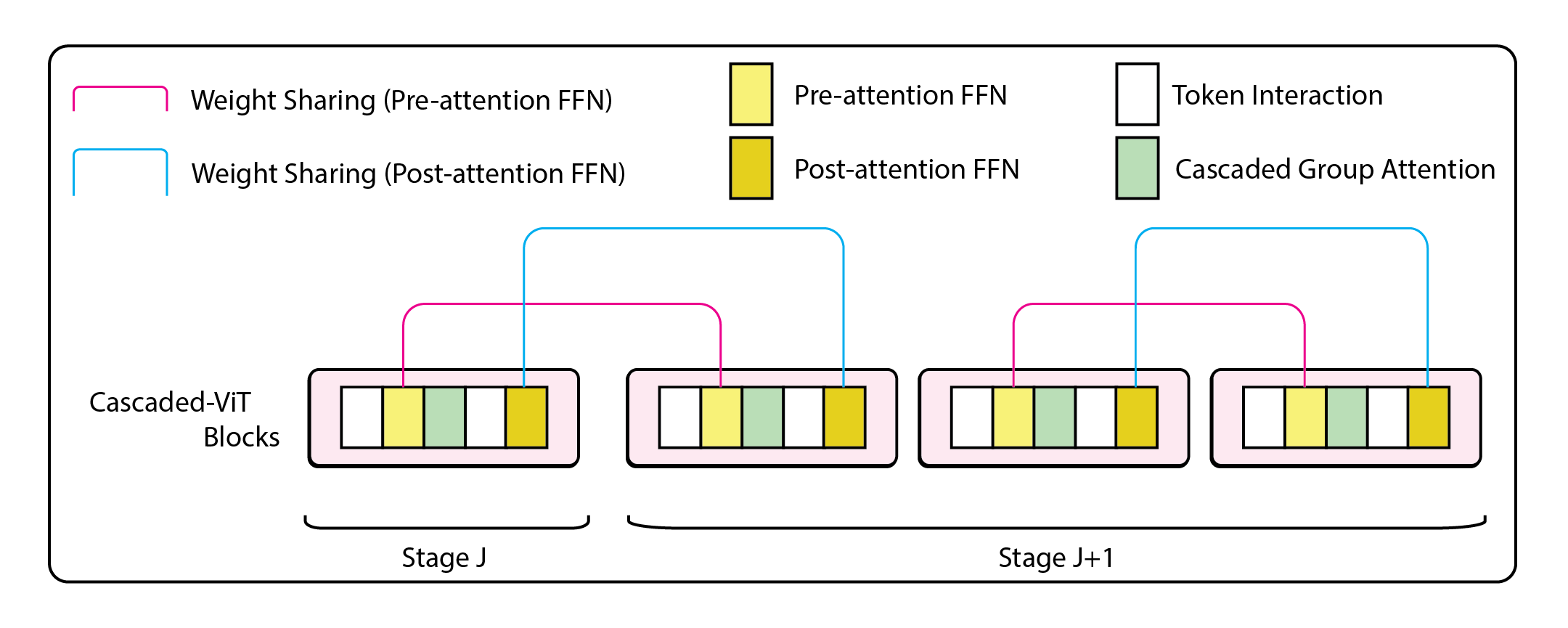}
    }
    \caption{An overview of weight sharing. Specifically, the pre-attention FFN modules in successive Cascaded-ViT blocks share weights, and similarly, the post-attention FFN modules also share weights across blocks. Although Cascaded-ViT blocks are organized into different stages, the current implementation applies weight sharing across stage boundaries, i.e., between all successive blocks regardless of stage. When the total number of blocks is odd, the final block does not participate in weight sharing.}
    \label{fig:weight-sharing}
\end{figure}

\begin{table}[t]
\centering
\begin{adjustbox}{width=0.7\columnwidth, center}
    \begin{tabular}{l | c | c | c | c  c  c  c  c}
        \toprule
        \multirow{1}{*}{\textbf{Model}} &
        \multirow{1}{*}{\textbf{No.}} & 
        \multicolumn{2}{c|}{\textbf{Ablation}} &
        \multirow{1}{*}{\textbf{Epochs}} & 
        \multirow{1}{*}{\makecell[c]{\textbf{Top-1}\\(\%)}} &
        \multirow{1}{*}{\makecell[c]{\textbf{Params}\\(M)}} &
        \multirow{1}{*}{\makecell[c]{\textbf{FLOPs}\\(M)}} &
        \multirow{1}{*}{\makecell[c]{\textbf{Energy/img}\\(mJ)}} \\
        \cmidrule{3-4}
        & & CCFFN & Wt. Shr. & & & & \\
        \midrule
        \rowcolor{gray!20}
        \multirow{2}{*}{\cellcolor{white}CViT-S} 
        & 0 & \cmark & \xmark & 300 & 59.5 & 1.9 & 67 & $471\pm29$ \\[1pt]
        & 1 & \xmark & \cmark & 300 & 58.5 & 2.2 & 88 & $506\pm10$ \\[1pt]
        \midrule
        & & No. Chunks & Exp. Ratio & \\
        \cmidrule{2-9}
        \rowcolor{gray!20}
         \multirow{6}{*}{\cellcolor{white}CViT-M} 
        & 2 & 2 & 2.5 & 80 & 62.83 & 3.5 & 173 & $571\pm13$ \\[1pt]
        & 3 & 2 & 4  & 80 & 63.80  & 4.2 & 201 & $591\pm19$ \\[1pt]
        & 4 & 4 & 2.5 & 80 & 60.80  & 2.9 & 150 & $555\pm13$ \\[1pt]
        & 5 & 4 & 4 & 80 & 61.54 & 3.2 & 164 & $561\pm16$ \\[1pt]
        \cmidrule{2-9}
        & & Cascade & Project & \\
        \cmidrule{2-9}
        & 6 & \cmark & \cmark & 80 & 58.7 & 3.9 & 192 & $581\pm19$ \\[1pt]
        & 7 & \xmark & \xmark & 80 & 62.60 & 3.5 & 173 & $578\pm16$ \\[1pt]
        \bottomrule
      \end{tabular}
  \end{adjustbox}
\caption{Ablation of our Cascaded-ChunkFFN module and design choices for CViT-M. Ablation No. 1 is performed on CViT-S. Wt. Shr. means clustered weight sharing.}
  \label{tbl:ablation}
\end{table}

\subsection{Ablation}  
\label{sec:ablation}

We conduct an ablation study to investigate the impact of various design elements within the CCFFN module on the performance of our CViT architecture using the ImageNet-1K dataset~\cite{5206848}. Each experiment involves training CViT-M for 80 epochs to balance result maturity with computational efficiency. The sole exception is Experiment~1, where we use CViT-S and train for 300 epochs to ensure that performance differences are clearly distinguishable and representative.

We begin by evaluating the effectiveness of the CCFFN module as a compression mechanism by replacing it with a standard weight-sharing FFN and observing its effect on accuracy and parameter count. Subsequently, we analyze the influence of two key hyperparameters: the number of chunks and the expansion ratio within the Chunk-FFN submodules. Finally, we assess the utility of cascading and projecting intermediate outputs within the CCFFN structure.

The results of these ablation experiments are summarized in \hyperref[sec:ablation]{Table~\ref{tbl:ablation}}, providing empirical insights into the design trade-offs and guiding principles for building efficient transformer-based models under resource constraints.

\textbf{Effectiveness of CCFFN.} Experiment~1 evaluates a clustered weight-sharing mechanism as an alternative to the proposed CCFFN module. Since full-scale weight sharing can severely constrain model flexibility and degrade performance~\cite{zhang2020deeper}, we adopt a more moderate strategy by grouping FFNs in pairs and sharing weights within each pair, as suggested in~\cite{zhang2020deeper}. 
To enhance this setup, the expansion ratio of the final post-attention FFN layer is increased from 2$\times$ to 4$\times$. However, this configuration leads to a substantial increase in both FLOPs and parameter count, with energy consumption rising by 7.6\%. In contrast, the CCFFN module achieves a 1\% higher Top-1 accuracy while significantly reducing FLOPs, parameter count, and energy usage. For this experiment, CCFFN used a configuration of two chunks and a 2.5$\times$ expansion ratio, selected based on results from subsequent experiments.

\textbf{Optimal Number of Chunks and Expansion Ratio.} Experiments~2 to~5 explore various combinations of chunk count and expansion ratios to identify the best trade-off between accuracy, model size, and energy efficiency. The configuration with two chunks and a 2.5$\times$ expansion ratio achieves the most favorable balance and is adopted as the default across our model family. Increasing the expansion ratio to 4$\times$ (Experiment~3) yields a marginal $\sim$1\% accuracy gain but comes at a significant cost in parameters, FLOPs, and energy. Conversely, increasing the number of chunks to four (Experiments~4 and~5) reduces computational cost and energy usage but results in a notable drop in classification accuracy. These results suggest that the two-chunk, 2.5$\times$ configuration provides the best overall efficiency.

\textbf{Impact of Cascading and Projection.} Experiments~6 and~7 examine the benefits of cascading and projecting intermediate outputs in the CCFFN structure using the previously selected efficient configuration. When projection is applied, the model experiences a steep decline in accuracy, likely due to insufficient gradient flow in the projection branch. Removing cascading altogether, and instead processing FFNs in parallel, results in a 0.23\% accuracy drop and increased energy consumption. These results confirm that the cascading mechanism contributes positively to both performance and energy efficiency.

\section{Conclusion}
In this paper, we introduced a novel Vision Transformer architecture, \textbf{CViT}, which enhances the efficiency of Feed Forward Networks (FFNs) through the proposed Cascaded ChunkFFN (CCFFN) modules. Our model achieves competitive Top-1 classification accuracy on the ImageNet-1K dataset~\cite{5206848} while offering significant reductions in parameter count, FLOPs, and energy consumption—making it well-suited for deployment in resource-constrained environments.
Despite its strengths, CViT exhibits certain limitations. Specifically, the initialization of multiple small FFNs within CCFFN can introduce latency overhead on GPUs due to increased kernel launches. Additionally, the division of the input feature map into smaller chunks may lead to a slight reduction in model capacity, potentially impacting accuracy.
As part of future work, we aim to improve the classification performance of CViT while preserving its compute and energy efficiency. This includes exploring more expressive chunk processing strategies and adaptive chunking mechanisms, as well as incorporating lightweight attention modules to enhance model capacity without significantly increasing overhead.

\bibliographystyle{unsrt}  
\bibliography{main}

\end{document}